\documentclass[letterpaper, 10 pt, conference]{ieeeconf}  

\IEEEoverridecommandlockouts                              

\usepackage{xcolor}
\usepackage[linesnumbered,ruled,vlined]{algorithm2e}
\usepackage{graphicx}
\usepackage{multirow}

\usepackage{amsmath}
\usepackage{color}
\usepackage{amssymb}
\usepackage{booktabs}
\usepackage[colorlinks, linkcolor=red]{hyperref}
\usepackage{array}
\usepackage{float}
\SetKwInput{KwInput}{Input}                
\SetKwInput{KwOutput}{Output} 

\newcolumntype{C}[1]{>{\centering\arraybackslash}p{#1}}

\newcommand{\para}[1]{\vspace{.05in}\noindent\textbf{#1}}

\title{\LARGE \bf
	Towards Robust Part-aware Instance Segmentation
	 \\ for Industrial Bin Picking
}

\author{Yidan Feng$^{*}$, Biqi Yang$^{*}$, Xianzhi Li, Chi-Wing Fu, Rui Cao, Kai Chen, Qi Dou,\\
	 Mingqiang Wei, Yun-Hui Liu, and Pheng-Ann Heng
	\thanks{$^{*}$ denotes equal contribution.}
	\thanks{Y. Feng and M. Wei are with the School of Computer Science and Technology, Nanjing University of Aeronautics and Astronautics.
	X. Li is with the School of Computer Science and Technology, Huazhong University of Science and Technology. 
	Y. Feng, B. Yang, X. Li, C.-W. Fu, R. Cao, K. Chen, Q. Dou, and P.-A. Heng are with the Department of Computer Science and Engineering, The Chinese University of Hong Kong.
	Y.-H. Liu is with the Department of Mechanical and Automation Engineering, The Chinese University of Hong Kong.}%
	\thanks{Corresponding author: Xianzhi Li
		{(\tt\small lixianzhi123@gmail.com)}}%
}

\begin{document}

	\maketitle
	\thispagestyle{empty}
	\pagestyle{empty}

	\begin{abstract}
		Industrial bin picking is a challenging task that requires accurate and robust segmentation of individual object instances. Particularly, industrial objects can have irregular shapes, that is, thin and concave, whereas in bin-picking scenarios, objects are often closely packed with strong occlusion. 
		To address these challenges, we formulate a novel part-aware instance segmentation pipeline. The key idea is to decompose industrial objects into correlated approximate convex parts and enhance the object-level segmentation with part-level segmentation. We design a part-aware network to predict part masks and part-to-part offsets, followed by a part aggregation module to assemble the recognized parts into instances. To guide the network learning, we also propose an automatic label decoupling scheme to generate ground-truth part-level labels from instance-level labels.
		Finally, we contribute the first instance segmentation dataset, which contains a variety of  industrial objects that are thin and have non-trivial shapes. 
		%
		%
		%
		Extensive experimental results on various industrial objects demonstrate that our method can achieve the best segmentation results compared with the state-of-the-art approaches.
		
	\end{abstract}
	
	
	
	\section{INTRODUCTION}
	\label{sec:intro}
	
	Instance segmentation is a fundamental task to support industrial bin picking. It requires not only correct detection of objects but also precise segmentation of each object.
	The segmentation accuracy affects the quality of subsequent tasks such as 6D pose estimation and robotic grasping.
	
\begin{figure}[t]
		\centering
		\includegraphics[width=\linewidth]{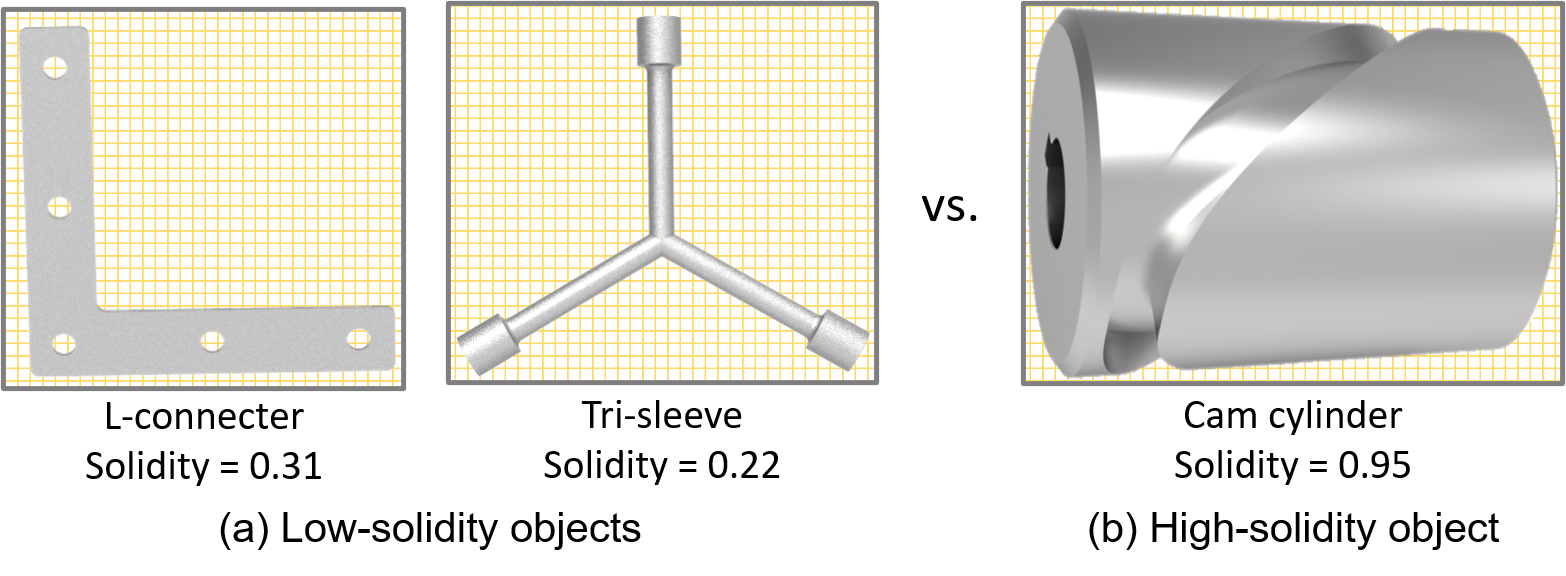}
		\vspace*{-8mm}
		\caption{Examples of low-solidity objects (a) and high-solidity object (b).}
		\label{fig:solidity}
		\vspace*{-1.5mm}
	\end{figure}

\begin{figure}[t]
		\centering
		\includegraphics[width=\linewidth]{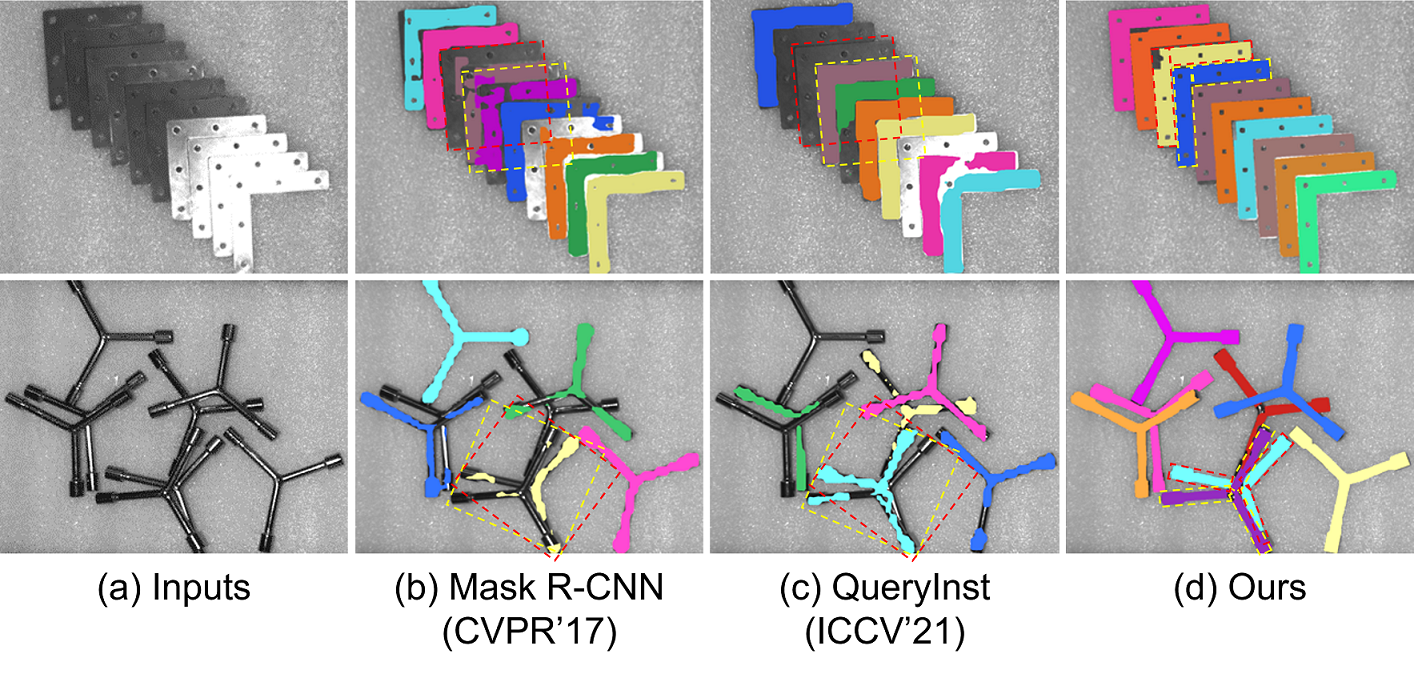}
		\vspace*{-8mm}
		\caption{Comparing the instance segmentation results produced by different methods (b-d) on two typical low-solidity industrial objects. See example bounding boxes in dashed lines of closely packed objects, highly overlapped boxes severely confuses object-level segmentation methods, while our method accurately segment all instances thanks to our part-level setting.
		}
		\label{fig:teaser}
		\vspace*{-5mm}
	\end{figure}

	Unlike everyday objects in daily environments, industrial objects can be thin and locally-concave in shape, which leads to low occupacy to its bounding box in camera view. To quantify such effects, we define the solidity of an object as 
	\begin{quote}
		\em{the minimum proportion of the object's bounding box occupied by the object mask.}
	\end{quote}
	%
	%
	%
	Figure~\ref{fig:solidity} shows example objects of low- and high-solidity.
	Also, industrial bin picking often suffers from close packing and severe occlusion, which brings additional challenges to obtain precise instance segmentation masks for 
	low-solidity objects.
	In bin picking, adjacent objects may be located very closely (see Figure~\ref{fig:teaser} (a)), so the bounding boxes of adjacent low-solidity objects can have high overlap, making it hard to accurately and robustly segment the object instances.
	
	Existing instance segmentation methods~\cite{he2017mask,chen2019tensormask,wang2020solov2,fang2021queryinst} are designed mostly for everyday objects.
	They cannot effectively handle industrial bin picking due to the close object packing with severe occlusion, especially for low-solidity objects; see Figure~\ref{fig:teaser} for results produced by other methods vs. ours.
	%
	Though some other methods~\cite{li2018learning,wada2019joint,follmann2019oriented,xie2021unseen} are specifically designed for bin picking, none of them explore segmenting low-solidity objects, like those in Figure~\ref{fig:solidity} (a),
	which are quite common in industrial bin picking. 
	
	In this work, we formulate a novel part-aware instance segmentation pipeline to segment industrial objects accurately and robustly under bin-picking scenarios.
	Our key idea is to automatically \emph{decompose a low-solidity object into multiple high-solidity parts} and enhance the instance segmentation task with part awareness.
	The core is a part-aware instance segmentation network that predicts both part masks and part-to-part offsets; then, we further feed them into our graph-based part aggregation module to obtain the final object instances.
	%
	Our network is guided by part-level labels instead of instance-level labels. Also, based on approximate concavity analysis, we propose an automatic label decoupling scheme to generate ground-truth part-level labels from instance-level labels.
	%
	%
	Our contributions are summarized as:
	\begin{itemize}
		\item To the best of our knowledge, we are the first to tackle robust instance segmentation for bin picking by developing a novel \emph{part-aware solution}. 
		\item To address the lack of data, we contribute the first instance segmentation dataset tailored for low-solidity industrial objects in bin picking.
		\item Extensive experiments demonstrate that our proposed network can accurately and robustly segment various industrial objects, outperforming the state-of-the-art approaches. See Figure~\ref{fig:teaser} and more results in Section~\ref{sec:experiment}.
	\end{itemize}

	\section{Related Work}
	\label{sec:rw}
	In this section, we first briefly review deep-learning-based instance segmentation works on natural images, and then discuss works that are tailored for bin-picking scenarios. 
	
	\para{Instance segmentation on natural images.} \ 
	Existing instance segmentation approaches designed for natural images can be grouped into the following two categories.
	
	%
	Single-stage methods, fueled by anchor-free object detection~\cite{duan2019centernet}, focus on designing various ways to represent or parameterize instance masks and to formulate networks to learn such parameterization.
	For example, some works  explicitly represent a mask by its geometrical contour~\cite{xu2019explicit,xie2020polarmask} or utilize structured 4D tensors~\cite{chen2019tensormask} to encode both object position and relative mask position; Others~\cite{bolya2019yolact,chen2020blendmask,lee2020centermask,wang2020solo,wang2020solov2} implicitly generate a set of prototype masks and assemble the final masks by learned coefficients or saliency.
	%
	%
	Generally, the above methods gain high accuracy when segmenting everyday objects, e.g., those in the COCO dataset~\cite{lin2014microsoft}.
	However, the parameterization of instance masks may not work well for industrial bin picking, especially for low-solidity objects; see Figure~\ref{fig:teaser} (c) for examples.
	
	
	Two-stage methods perform object detection and mask regression consecutively.
	Generally, these methods first detect objects by extracting a bounding box for each object instance, followed by predicting foreground mask inside each bounding box. 
	Mask R-CNN~\cite{he2017mask}, as a typical two-stage approach, introduces the region proposal network to locate candidate object bounding boxes, followed by the RoIAlign operation to extract features from each candidate box.
	Finally, the mask head generates the instance masks guided by the output of the classification head and box regression head. 
	Many Mask R-CNN variants~\cite{liu2018path,chen2018masklab,huang2019mask} have been proposed to further improve the mask head. 
	%
	Despite the promising performance achieved by these two-stage approaches, an obvious constraint is that each bounding box contains only one mask.
	Yet, in industrial bin picking, low-solidity objects can closely stack over one another with heavily-overlapped bounding boxes, thus leading to inaccuracy for the existing two-stage approaches; see Figure~\ref{fig:teaser} (b) as examples.
	Section~\ref{sec:experiment} presents more comparison results.
	\para{Segmentation for bin-picking scenarios.} \ 
	%
	Existing works mainly solve the instance segmentation under bin-picking settings from two perspectives: (i) towards high-quality segmentation given occluded scenes, and (ii) towards method generalization due to the limited bin-picking data and the gap between simulation data and real scenes.
	To address the occluded scenes, Zeng et al.~\cite{zeng2017multi} proposed to leverage multi-view RGB-D data, while Wada et al.~\cite{wada2019joint} explored the collaboration of semantic and instance segmentation.
	To improve method generalization, two works~\cite{schwarz2017data,schwarz2018rgb} were proposed to only fine-tune the pre-trained networks for the specific object domain towards stronger generalization ability.
	Further, Danielczuk et al.~\cite{danielczuk2019segmenting} created a photorealistic bin-picking dataset to aid unknown segmentation from real depth images. 
	Recently, Xie et al.~\cite{xie2020best} proposed to fuse depth map and RGB image for better segmenting on unseen objects. 
	Yet, the above methods mainly explore the segmentation of everyday objects for bin picking, not industrial objects.

	On the other hand, to segment industrial objects, Matsumura et al.~\cite{matsumura2019learning} first tried to grasp potentially tangled objects, but merely focuses on single object picking rather a solution for accurate segmentation. 
	Follmann et al.~\cite{follmann2019oriented} employed a rotated bounding box to segment thin industrial objects, but fail to handle challenging low-solidity objects. 
	To the best of our knowledge, this paper makes the first step for segmenting low-solidity instances cluttered in a bin. 
	\section{ILS Dataset}
	\label{sec:dataset}

	\begin{figure}[t]
		\centering
		\includegraphics[width=\linewidth]{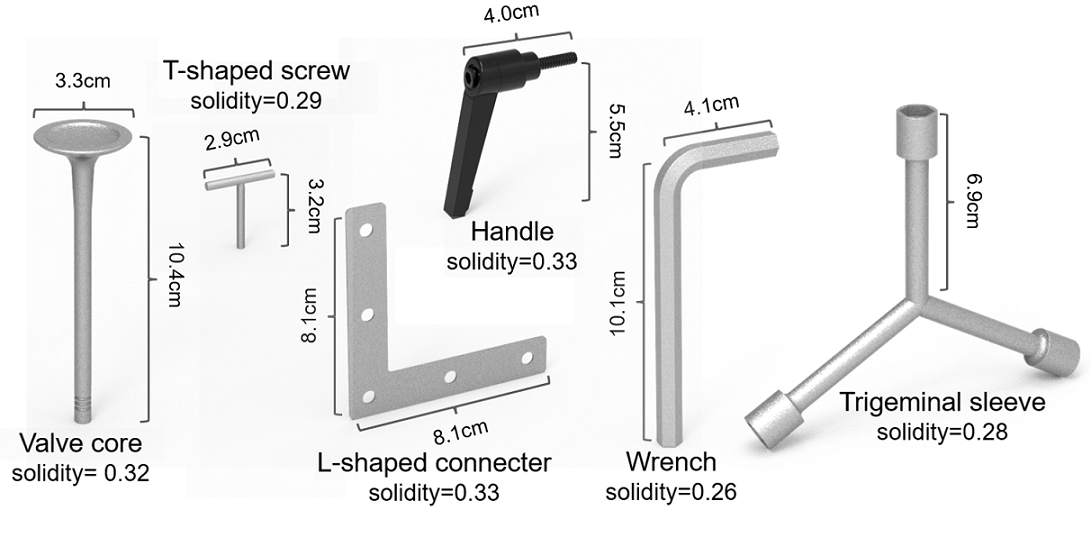}
		\vspace*{-9mm}
		\caption{An overview of the objects in our compiled dataset, each includes its object name, size, and solidity value.}
		\label{fig:model}
		\vspace*{-5mm}
	\end{figure}

	Considering that the objects provided in existing bin-picking datasets~\cite{calli2015benchmarking,kleeberger2019large} are mostly high-solidity objects,  
	we first contribute a new large-scale instance segmentation dataset with industrial low-solidity objects (referred to as ``ILS'') for industrial bin picking, which is detailed below. 
	
	\para{3D model description.} \ 
	In our ILS dataset, we collected six common low-solidity industrial objects with varying shapes and sizes, including T-screw, Wrench, Tri-sleeve, L-connecter, Valve and Handle, as shown in Figure~\ref{fig:model} with their corresponding sizes and solidity values.

	\begin{figure}[t]
		\centering
		\includegraphics[width=\linewidth]{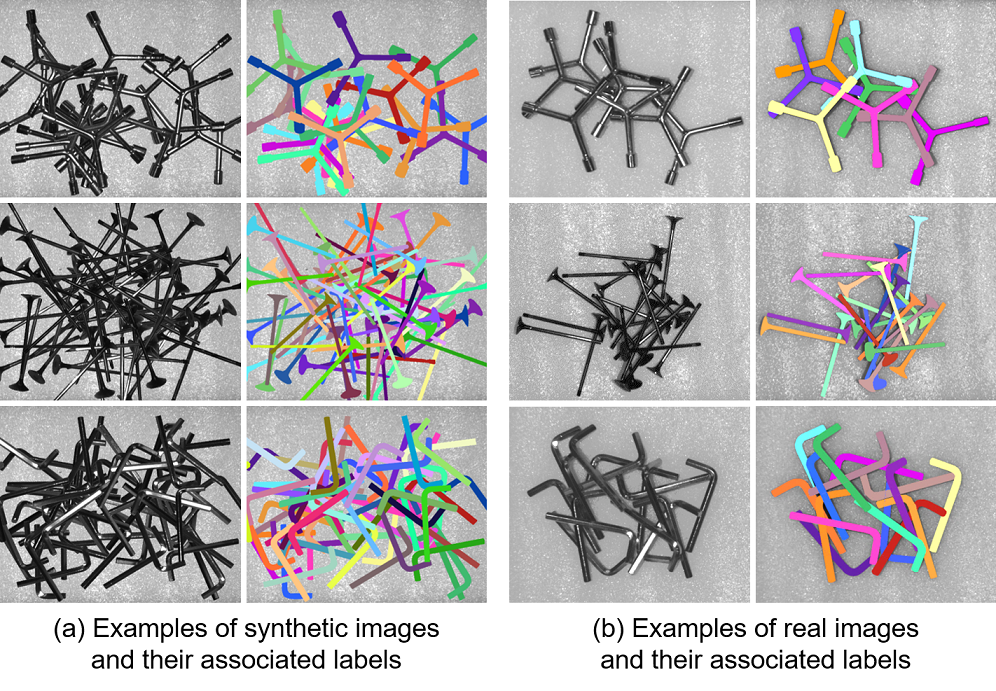}
		\vspace*{-8mm}
		\caption{Examples of our prepared synthetic data (left) and real data (right).}
		\label{fig:dataset}
		\vspace*{-3mm}
	\end{figure}

	\begin{figure*}[t]
		\centering
		\includegraphics[width=\textwidth]{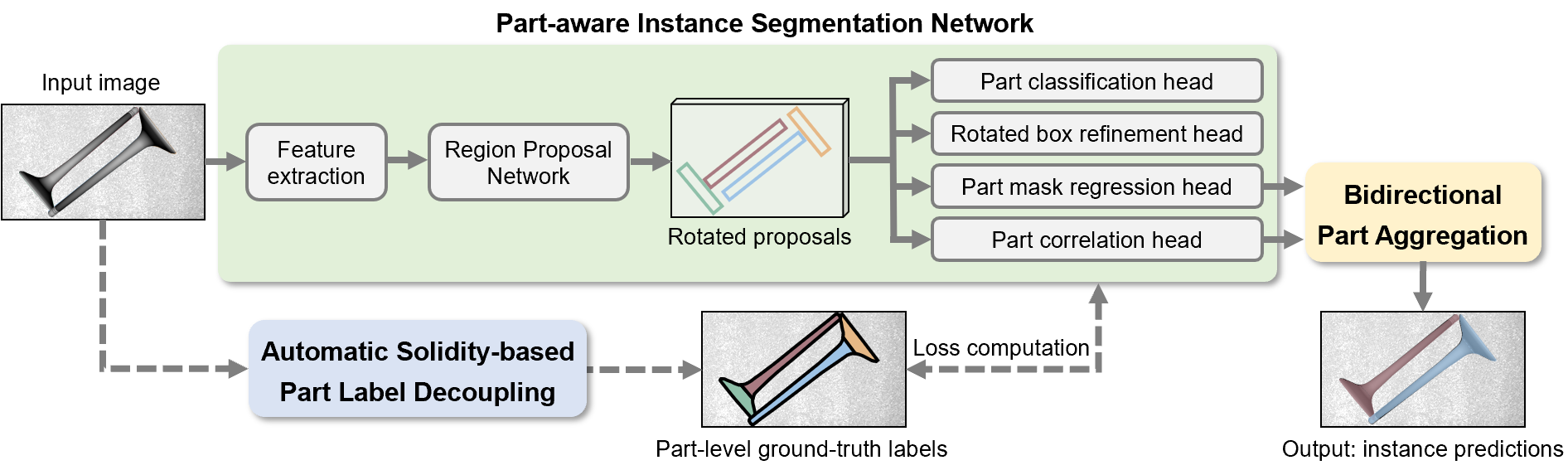}
		\vspace*{-8mm}
		\caption{Proposed pipeline. Given an input image, the part-aware instance segmentation network (green) first predicts part masks and part correlation, which will be fed into a bidirectional part aggregation module (yellow) to obtain the complete instances during the testing stage. To guide the network training, an automatic label decoupling scheme (blue) is introduced to generate part-level ground-truth labels from instance labels for loss computation.
		}
		\label{fig:pipeline}
		\vspace*{-3.5mm}
	\end{figure*}
	
	\para{Dataset description.} \ 
	In our dataset, we prepared synthetic data for training and real data with human labeling for testing.
	Specifically, we followed the copy-paste strategy \cite{ghiasi2021simple, dwibedi2017cut,dvornik2018modeling} to generate photo-realistic synthetic images with ground-truth instance labels. 
	The key idea of copy-paste strategy is to manually-crop completely exposed objects of varying poses from a small set of real-world images (less than 10) under different lighting conditions as templates, and then paste these templates on a real background many times with random rotations and positions. 
	To create a crowding effect, we randomly copied about 20-100 times in each scene; see Figure~\ref{fig:dataset} (a) for examples of our generated synthetic images and their associated ground-truth instance labels.
	Note that, such copy-paste data generation allows us to easily extend the scale of our training data at a low cost. 
	
	To collect real-world data, we mounted a Smarteye HD-1000 high-precision stereo camera above the bin to capture real images with a resolution of 1920$\times$1280. 
	Note that, most of the industrial cameras, including ours, can only capture grayscale images for cost-effective, and the depth images for metal objects are of very poor quality and thus are not used in our method.
	Each real image is hand-labeled using labelme \cite{labelme2016}; see Figure~\ref{fig:dataset} (b) for examples.
	
	Overall, for each object, we totally prepared 1,000 synthetic scenes (or images) and about 100 real scenes.
	All of these scenes have full per-pixel annotations.
	Please refer to Table~\ref{table:dataset} for the detailed statistics of our compiled dataset.

	
	\begin{table}[]
		\vspace*{1mm}
		\caption{Statistics of our compiled dataset.}
		\vspace*{-3.5mm}
		\centering
		\resizebox{1.0\linewidth}{!}{
			\begin{tabular}{@{\hspace{1mm}}C{1.9cm}@{\hspace{1mm}}|@{\hspace{1mm}}C{0.9cm}@{\hspace{1mm}}C{0.9cm}@{\hspace{1mm}}C{1.2cm}@{\hspace{1mm}}C{1.4cm}@{\hspace{1mm}}C{0.5cm}@{\hspace{1mm}}C{1.1cm}@{\hspace{1mm}}C{0.6cm}@{\hspace{1mm}}}
				\toprule[1pt]
				Objects & T-screw & Wrench &  Tri-sleeve & L-connector & Valve & Handle \\ \hline
				number of & \multirow{2}*{1000} & \multirow{2}*{1000} & \multirow{2}*{1000} & \multirow{2}*{1000} & \multirow{2}*{1000} &  \multirow{2}*{1000} \\
				synthetic scenes &  &  &  &  &  &  \\ \hline
				number of & \multirow{2}*{81} & \multirow{2}*{104} & \multirow{2}*{100} & \multirow{2}*{98} & \multirow{2}*{120} & \multirow{2}*{111} \\
				real scenes &  &  &  &  &  &  \\
				\bottomrule[1pt]
			\end{tabular}
		}
		\vspace*{-7mm}
		\label{table:dataset}
	\end{table}
	
	
	\section{Method}
	\label{sec:method}
	\subsection{Overview}
	\label{subsec:overview}
	As discussed in Section~\ref{sec:intro}, the key observation of low-solidity objects is that, the instance mask area is much smaller than its associated bounding box area, so these objects are often thin and locally concave.
	Hence, two closely-located objects in bin-picking scenes may share almost the same bounding box, thus leading to the confusion of existing segmentation works.
	To enable an accurate segmentation of low-solidity objects, different from previous works that treat each instance as a whole, we propose to decompose the low-solidity object into multiple high-solidity parts, and design our instance segmentation network to be part-aware.
	
	Figure~\ref{fig:pipeline} depicts the pipeline of our method.
	Generally, given an input image, we first design a part-aware instance segmentation network with multiple heads (Section~\ref{subsec:network}) to predict both part masks and part correlations.
	During training, to guide the network learning, we propose an automatic label decoupling scheme (Section~\ref{subsec:label_decouple}) to transform the original instance-level ground-truth instance labels into part-level labels.
	During testing, we feed the predicted part masks and correlations into a bidirectional part aggregation module (Section~\ref{subsec:aggregation}) to obtain the final object-level instance masks.
	
	
	\subsection{Part-aware Instance Segmentation Network}
	\label{subsec:network}
	
	
	The green box in the middle of Figure~\ref{fig:pipeline} presents the architecture of our part-aware instance segmentation network.
	%
	Specifically, inspired by Mask R-CNN \cite{he2017mask}, we first feed the input image into a feature embedding module to extract image-level features. 
	Next, in Mask R-CNN, a region proposal network is used to select multiple proposals (i.e., candidate instance masks) that are represented by bounding boxes from the extracted features.
	However, in our task, the industrial objects are thin and locally concave.
	Hence, a regular bounding box tends to contain too much useless background information.
	To enable the extraction of tight and distinctive local representations for the potential parts, we here choose to select the \emph{rotated} proposals from a set of template anchor boxes produced by enumerating box sizes, aspect ratios, and angles.
	Each proposal is represented by a rotated bounding box $(x,y,h,w,a)$, encoding box center coordinates $(x,y)$, height $h$, width $w$, and the box orientation $a$ to the major axis.
	After that, we then follow Mask R-CNN to extract the associated features of each proposal and design four heads to regress part-level predictions, including:
	%
	
	\para{(1) Part classification head} to predict the possibility that each proposal contains a part, which is used to filter out redundant proposals for mask regression.
	
	\para{(2) Rotated box refinement head} to conduct multi-variable regression to further refine the rotated bounding box. 
	
	\para{(3) Part mask regression head} to regress the part mask by using fully convolutional layers.
	
	\para{(4) Part correlation head} to regress part-to-part correlations. In detail, for each reference part, this head regresses (i) the offset $\mathbf{u} \in \mathbb{R}^2$ from the center of the part's visible area to the actual center of the whole part; hence, if the reference part is totally visible without occlusion, then its $\mathbf{u}$ should be $(0, 0)$; 
	and also regresses (ii) the offset $\{\mathbf{v}\} \in \mathbb{R}^{(N-1)\times2}$ from the center of the reference part to the centers of the instance's other parts, where $N$ is the total number of parts in this instance.
	Intuitively, $\mathbf{u}$ helps correct the part center against occlusions, while $\{\mathbf{v}\}$ describes the correlation of each part to other parts, which encodes the intrinsic geometric structure of an object, thus being invariant against varying textures.
	%
	
	In the training phase, our network has a multi-task loss on each proposal as $\mathcal{L}=\mathcal{L}_{\text{cls}}+\mathcal{L}_{\text{mask}}+\mathcal{L}_{\text{rbox}}+\mathcal{L}_{\text{offset}}$, where  $\mathcal{L}_{\text{cls}}$, $\mathcal{L}_{\text{rbox}}$ and $\mathcal{L}_{\text{mask}}$ follow the typical losses in \cite{he2017mask}, and $\mathcal{L}_{\text{offset}}$ is defined by the smooth $\mathcal{L}_{1}$ loss:
	\begin{equation}
		\mathcal{L}_{\text {offset }}(\mathbf{o}, \mathbf{\hat{o}})=\sum_{i \in\{x, y\}} \begin{cases}0.5\left(o_{i}-\hat{o}_{i}\right)^{2}, & \text { if }\left|o_{i}-\hat{o}_{i}\right|<1 \\ \left|o_{i}-\hat{o}_{i}\right|-0.5, & \text { otherwise }\end{cases}
	\end{equation}
	where $\mathbf{o}=[\mathbf{u}, \{\mathbf{v}\}]$ denotes the predicted offsets for each part, and $\mathbf{\hat{o}}$ denotes the ground-truth offsets generated by our automatic solidity-based label decoupling (Section~\ref{subsec:label_decouple}).
	During testing, we feed the predicted part masks and their associated offsets into our bidirectional part aggregation module (Section~\ref{subsec:aggregation}) to obtain the complete instances. 
	\subsection{Automatic Solidity-based Label Decoupling}
	\label{subsec:label_decouple}
	%
	The goal of label decoupling is to automatically decouple an instance with ground-truth label $\hat{I}$ into $N$ high-solidity parts with labels $\{\hat{P}_i\}_{i=1}^N$, where $\hat{P}_i=[\hat{M}_i,\hat{\mathbf{o}}_i]$ contains the ground-truth binary part mask ${\hat{M}}_i$ and correlation offsets $\hat{\mathbf{o}}_i=[\hat{\mathbf{u}}_i,\{\hat{\mathbf{v}}\}_i]$ for the $i$-th part. 
	Note that, the part number $N$ is not a fixed pre-defined value, but dynamically determined by the geometric property of this instance.
	%
	
	\para{Part masks generation.} \ 
	We decompose an object instance   to generate approximate convex sub-masks that can be tightly enclosed by bounding boxes. 
	Specifically, given a binary mask ${M}$ that needs to be split, and the discretized contour points $\mathbf{C}=\{c\}$,
	we then follow \cite{lien2006approximate} to define the concavity of each point $c$ as $\operatorname{concave}(c)=\operatorname{dist}(c, \beta)$,
	where $\beta$ is the convex hull edge corresponding to $c$, and $\operatorname{dist}(c, \beta)$ is the Euclidean distance from $c$ to line $\beta$.
	Thus, the global concavity of a mask $M$ can be defined as:
	\begin{equation}
		\operatorname{concave}({M})=\max\{\operatorname{concave}(c)\}, \forall{c} \in \mathbf{C}.
	\end{equation}
	
\begin{figure}
		\centering
		\includegraphics[width=0.95\linewidth]{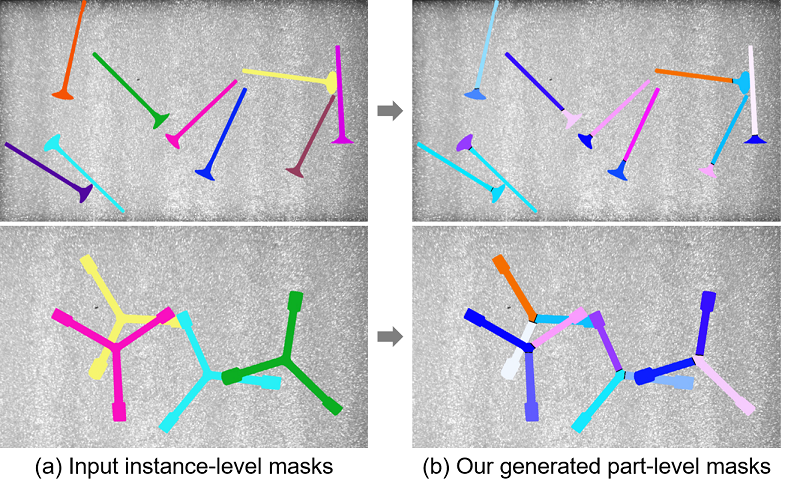}
		\vspace*{-4mm}
		\caption{Given the input scenes with ground-truth instance labels (a), our label decoupling algorithm can automatically generate ground-truth part masks (b) for network training.}
		\label{fig:concavity}
		\vspace*{-5mm}
	\end{figure}

	Based on this definition, we 
	recursively split the input mask, which is initialized as an unoccluded version of $\hat{I}$, into two sub-masks $\{{{M}_1},{{M}_2}\}$ until each sub-mask ${{M}_i}$ satisfies $\operatorname{concave}({{M}_i})<\tau$, where $\tau$ is a pre-defined parameter denoting the non-concavity tolerance. In each recursion, the mask is split through a cut line, which is defined by two key points: the start point
	$p_s=\arg \max\limits_{c\in \mathbf{C}}\{\operatorname{concave}(c)\}$ with maximum concavity, and the endpoint $p_e\in \mathbf{C}$ that (i) divides the mask into two individual parts and (ii) reaches a proper balance between minimizing $\operatorname{dist}(p_s,p_e)$ and maximizing $\operatorname{concavity}(p_e)$.
	After collecting $\{M_i\}$, each ground-truth part mask $\hat{M}_i$ is obtained by intersecting ${{M}}_i$ with ground-truth instance mask $\hat I$. Figure \ref{fig:concavity} presents the decoupling results, showing that given instance masks, the above proposed algorithm can effectively generate high-solidity part masks.

	\para{Part offsets generation.} \ 
	Given a decoupled part mask ${M}_i$, the ground-truth offset $\{\mathbf{{\hat{v}}}\}_i$ relative to other remaining parts of this instance can be calculated as: 
	\begin{equation}
		\label{equ:v}
		\{\mathbf{{\hat{v}}}\}_i = \{\phi_j - \phi_i \mid j \in [1,N], j \neq i\}
	\end{equation}
	where $\phi$ denotes the part center, which is calculated as the average of all pixel coordinates of the associated part.
	The predicted $\{\mathbf{v}\}_i$ will construct correlations between unoccluded parts. However, the true predicted parts always suffer from strong occlusion and overlapping. Hence we collect another ground-truth offset $\mathbf{\hat{u}}_i=\phi_i-\hat{\phi}_i$. During testing, the predicted $\mathbf{u}_i$ will move the center of a predicted occluded mask to its complete mask. After correcting centers, we can utilize $\{\mathbf{v}\}_i$ to aggregate the predicted parts into object instances.

	%
	%
	%
	%
	%
	%
	%
	%
	%
	
	\begin{table*}[t]
		\vspace*{-2mm}
		\caption{Comparing the instance segmentation quality of state-of-the-art methods and ours on the ILS dataset.}
		\vspace*{-3mm}
		\centering
		\resizebox{1.0\linewidth}{!}{
			\begin{tabular}{@{}c|ccc|cccccccccccccccccc@{}}
				\toprule
				\multirow{2}{*}{Method} 
				& \multicolumn{3}{c|}{Average}           & \multicolumn{3}{c}{Valve}                     & \multicolumn{3}{c}{Wrench}                    & \multicolumn{3}{c}{L-connecter}               & \multicolumn{3}{c}{T-screw}          & \multicolumn{3}{c}{Handle}                    & \multicolumn{3}{c}{Tri-sleeve}                \\ \cmidrule(l){2-22} 
				& $AP_{50}$   & $AP_{75}$   & mIoU  & $AP_{50}$          & $AP_{75}$          & mIoU          & $AP_{50}$           & $AP_{75}$         & mIoU          & $AP_{50}$          &$AP_{75}$         & mIoU          & $AP_{50}$          & $AP_{75}$          & mIoU          & $AP_{50}$           & $AP_{75}$         & mIoU                  & $AP_{50}$           & $AP_{75}$         & mIoU          \\ \midrule
				MaskRCNN                &  0.86      &  0.56      &  0.72     & 0.81     &  0.50             &  0.68             & 0.86          & 0.64          & 0.73          & 0.89          & 0.72          & 0.76          & 0.92          & 0.62          & 0.74          & 0.78          & 0.64          & 0.73          & 0.91          & 0.28          & 0.68          \\ 
				TensorMask              & 0.71       &0.33        &  0.62     & 0.45             &  0.11            &  0.47            &0.70            & 0.35             & 0.62             & 0.90          & 0.74          & 0.78 & 0.71          & 0.24          & 0.59          & 0.52          & 0.30          & 0.57                   & \textbf{0.98} & 0.29          & 0.71          \\
				SOLOv2                  & 0.82       &  0.64      & 0.71      &  0.75     & 0.56      &   0.67            & 0.76              & 0.66              &  0.68             & 0.77          & 0.64          &  0.71         & 0.92          & 0.71          &  0.75             &  0.92             & 0.87              &   0.83                  &  0.79             &0.37               &  0.63             \\
				QueryInst               &0.78        &0.41       & 0.64      & 0.93            &  0.52             &   0.72            & 0.68      &  0.35          &  0.60    & 0.86         &  0.65          &  0.73       &        0.55       &0.31              &  0.53           &   0.82           & 0.45            & 0.68               &   0.83            & 0.18              &   0.62            \\	
				\midrule
				
				Ours   &  \textbf{0.98}      &  \textbf{0.86}      & \textbf{0.85}      & \textbf{0.99} & \textbf{0.96} & \textbf{0.85} & \textbf{0.99} & \textbf{0.97} & \textbf{0.89} & \textbf{0.98} & \textbf{0.98} & \textbf{0.90} & \textbf{0.95} & \textbf{0.68} & \textbf{0.80} & \textbf{0.97} & \textbf{0.96} & \textbf{0.88}    & \textbf{0.98} & \textbf{0.69} & \textbf{0.77} \\ \bottomrule
		\end{tabular}}
		\label{tab:comparison}
		\vspace*{-3mm}
	\end{table*}
	
	\subsection{Bidirectional Part Aggregation}
	\label{subsec:aggregation}
	During inference, the part-level predictions are aggregated into complete object instances. Each part prediction has $N-1$ $v$ offsets, where each $v$ offset points to an associated part that belongs to the same instance. Ideally, the part predictions can be formed as complete correlation graphs, where the center of each part mask is a vertex and their part-to-part offsets $\{\mathbf{v}\}$ are the edges. The aggregation algorithm traverses each unvisited vertex and collects all its neighbors as an instance, while the network is expected to output precise masks and offsets of each part, thus enabling one-to-one part matching.  
	
	However, considering the approximation uncertainty, the actually matching can be modeled as a one-to-many problem, and constrained by the bidirectional property in the instance complete graph for accurate part assembly.
	Specifically, Given a random part mask $M_i$, for each of its valid predicted correlation offset $\mathbf{v} \in \{\mathbf{v}\}_i$, we collect its candidate pool, where each candidate part $M_j$ satisfies $\epsilon$-approximate matching:
	\begin{equation}
		\label{equ:epsilon}
		\| \overrightarrow{\phi_j-\phi_i}-\mathbf{v} \| \leq \epsilon, \mathbf{v} \in \{\mathbf{v}\}_i.
	\end{equation}
	Thus, the true sibling part index is obtained by:
	\begin{equation}
		j^*=\arg \min\limits_{j}\|\overrightarrow{\phi_i-\phi_j}-\mathbf{v}\|, \mathbf{v} \in \{\mathbf{v}\}_j.
	\end{equation}
	
	In this way, after a whole object instance is aggregated, we delete the associate parts from the predictions and continue to aggregate the next instance until no pairing meets Eq.~\eqref{equ:epsilon}.
	Note that, the predicted parts that fail to be paired will be filtered out, which benefits the removal of false positives since they are isolated from any part in a graph.

	\section{Experiments and Results}
	
	\label{sec:experiment}
	\subsection{Implementation Details}
	We implemented our network based on Detectron2 \cite{wu2019detectron2}.
	We used Adam optimizer with a learning rate of $5e-3$ and weight decay of $1e-4$ for training. 
	The non-concavity tolerance $\tau$ is set as $0.2*d_\text{short}$ in all experiments, where $d_\text{short}$ is the length of the shorter edge of the rotated bounding box. 
	During testing, after aggregating the complete instances, 
	we adopt morphological operations to refine the instance mask: the aggregated mask is first dilated to eliminate the narrow gaps between sibling parts, and then eroded to recover its original size. 
	On average, it takes $\sim$12 hours to train our network and $\sim$0.5s to segment an input scene with about 20 instances on one NVIDIA Titan Xp GPU.
	The picking experiments are based on 6D poses by adopting DenseFusion \cite{wang2019densefusion} based on our segmentation results.

	\begin{figure*}
		\centering
		\includegraphics[width=1.0\textwidth]{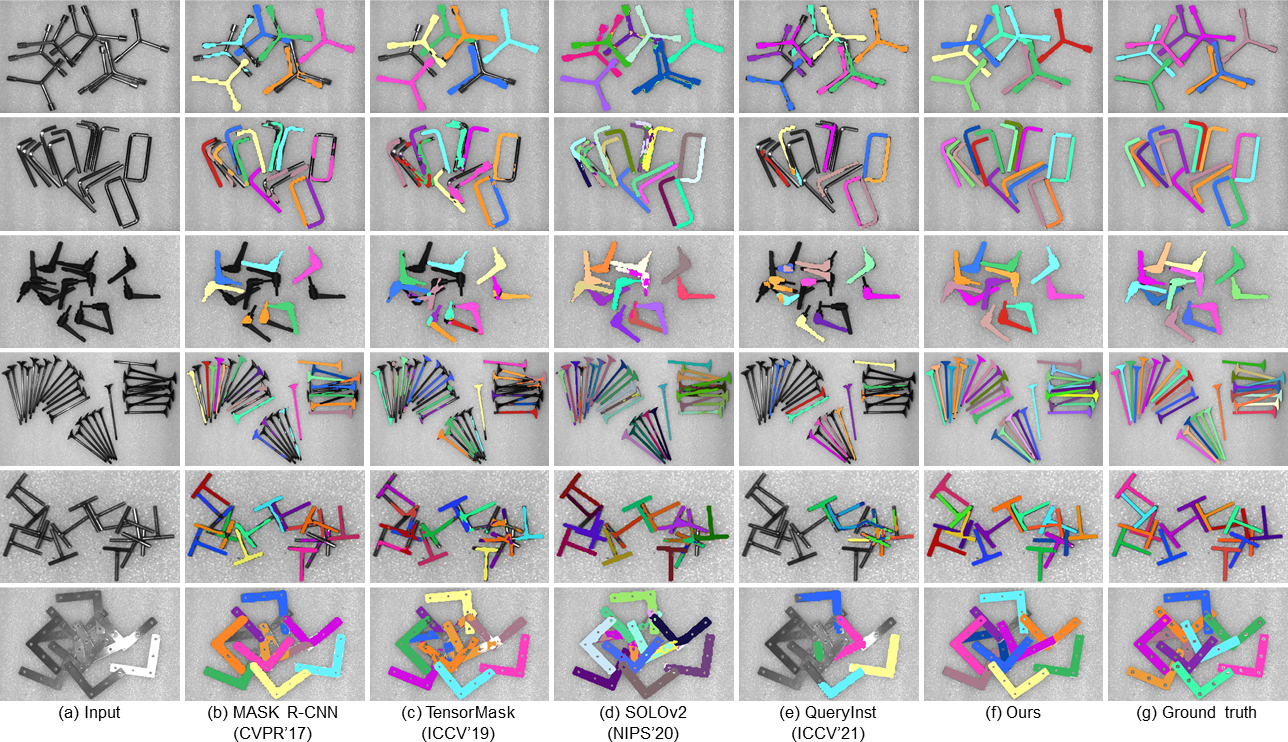}
		\vspace*{-6mm}
		\caption{Comparing the instance segmentation results produced by different methods (b-f) against ground truth (g). Our method consistently outperforms others with the most accurate segmentation results for all the test objects.}
		\label{fig:comparison}
		\vspace*{-3mm}
	\end{figure*}
	
	\subsection{Comparison with State-of-the-art Methods}
	We first compare our method with recent state-of-the-art single-stage methods, including SOLOv2 \cite{wang2020solov2}, QueryInst \cite{fang2021queryinst}, TensorMask \cite{chen2019tensormask}, as well as the two-stage method Mask R-CNN \cite{he2017mask}. 
	All comparison methods are re-trained on our ILS dataset using their provided codes and default configurations. 
	We follow \cite{he2017mask} to use average precision (AP) as the evaluation metric, where a predicted mask is considered as true positive if the IoU (intersection-over-union) between the ground truth and the predicted mask is larger than a predefined threshold, which is set to be $50\%$ and $75\%$ following \cite{he2017mask}. 
	Besides, we also report the mean IoU (mIoU) for comparing the segmentation quality.
	Table \ref{tab:comparison} shows the comparison results.
	Clearly, our method achieves the highest values on all the metrics over all the low-solidity objects.
	Particularly, our method yields nearly $100\%$ correct predictions for all the objects under $\text{AP}_{50}$. 
	The visual comparisons are presented in Figure \ref{fig:comparison}.
	Compared with existing works (b-e), our method (f) can accurately segment almost all the instances.
	
	\begin{figure}
		\centering
		\includegraphics[width=\linewidth]{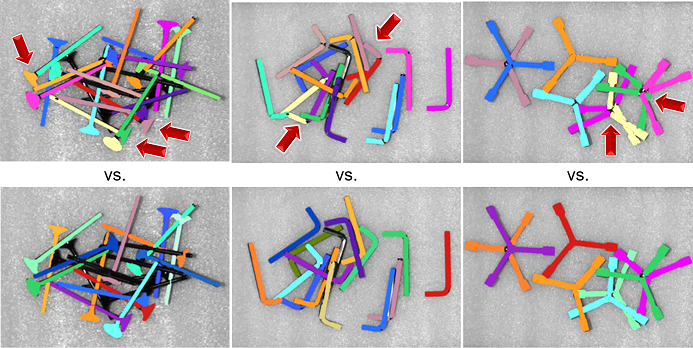}
		\vspace*{-7mm}
		\caption{Comparing part aggregation results by using the distance-based Hungarian method (top) and our bidirectional aggregation method (bottom).} 
		\label{fig:assembly}
		\vspace*{-4mm}
	\end{figure}
	
	\subsection{Analysis on Part Aggregation}
	To aggregate multiple parts into an instance, we design our network to learn part correlations, followed by our bidirectional part aggregation module.
	To verify the effectiveness of the above procedure, 
	we compare it with a traditional baseline method: distance-based Hungarian part aggregation.
	
	In detail, given two parts $M_i$ and $M_j$, their correlation weight can be defined as  $\operatorname{W}(M_i,M_j) $$=$$ \operatorname{dist}(M_i,M_j) + \lambda*\operatorname{IoU}(M_i,M_j)$,
	where $\operatorname{dist}(\cdot)$ is the Euclidean distance between two parts' centers, and $\operatorname{IoU}(\cdot)$ measures the overlapping degree. 
	Hence, this baseline method regards the optimal matching part of $M_i$ as the one with the highest $\operatorname{W}$, where the computational complexity is $\mathcal{O}(n^3)$.
	
	Figure \ref{fig:assembly} shows the comparison results by using the baseline aggregation method (top) and our proposed method (bottom).
	It can be observed that in the top row, the confusion inevitably happens in cluttered scenes when two parts are adjacent but belong to different instances; see the regions pointed by red arrows. In contrast, our method can accurately assemble multiple parts into an instance, since we directly infer the best pair for each part using a neural network, 
	thus being robust against confusing poses of cluttered objects. 
	Quantitatively, the average $\text{AP}_{50}$ produced by baseline method and our method is $0.78$ vs. $0.95$.
	Particularly, the computation complexity of our aggregation module is only $\mathcal{O}(n)$, as compared with $\mathcal{O}(n^3)$ of the baseline method.
	
	\begin{figure}
		\centering
		\includegraphics[width=1.0\linewidth]{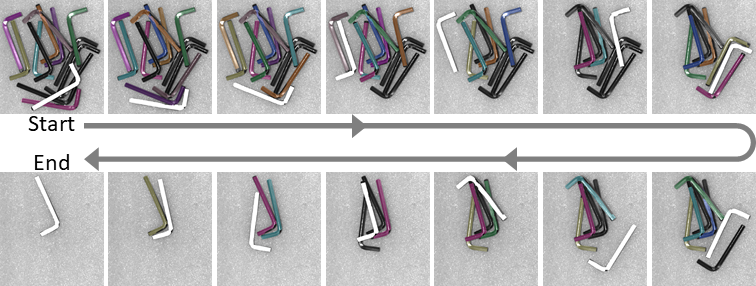}  
		\vspace*{-6mm}
		\caption{Our method consistently generates precise instance masks to support robot grasping. The object marked in white is the one to be grasped.}
		\label{fig:grasping}
		\vspace*{-5mm}
	\end{figure}

	\subsection{Robustness Analysis}
	\para{Continuous segmentation}: First, we conduct a real grasping experiment by continuously segmenting and picking objects one by one; see Figure~\ref{fig:grasping} for the segmentation results on sequential scenes, where the instance marked in white indicates the object to be grasped.
	Our method consistently generates precise instance masks to support robot grasping.
	
\begin{figure}
		\centering
		\includegraphics[width=\linewidth]{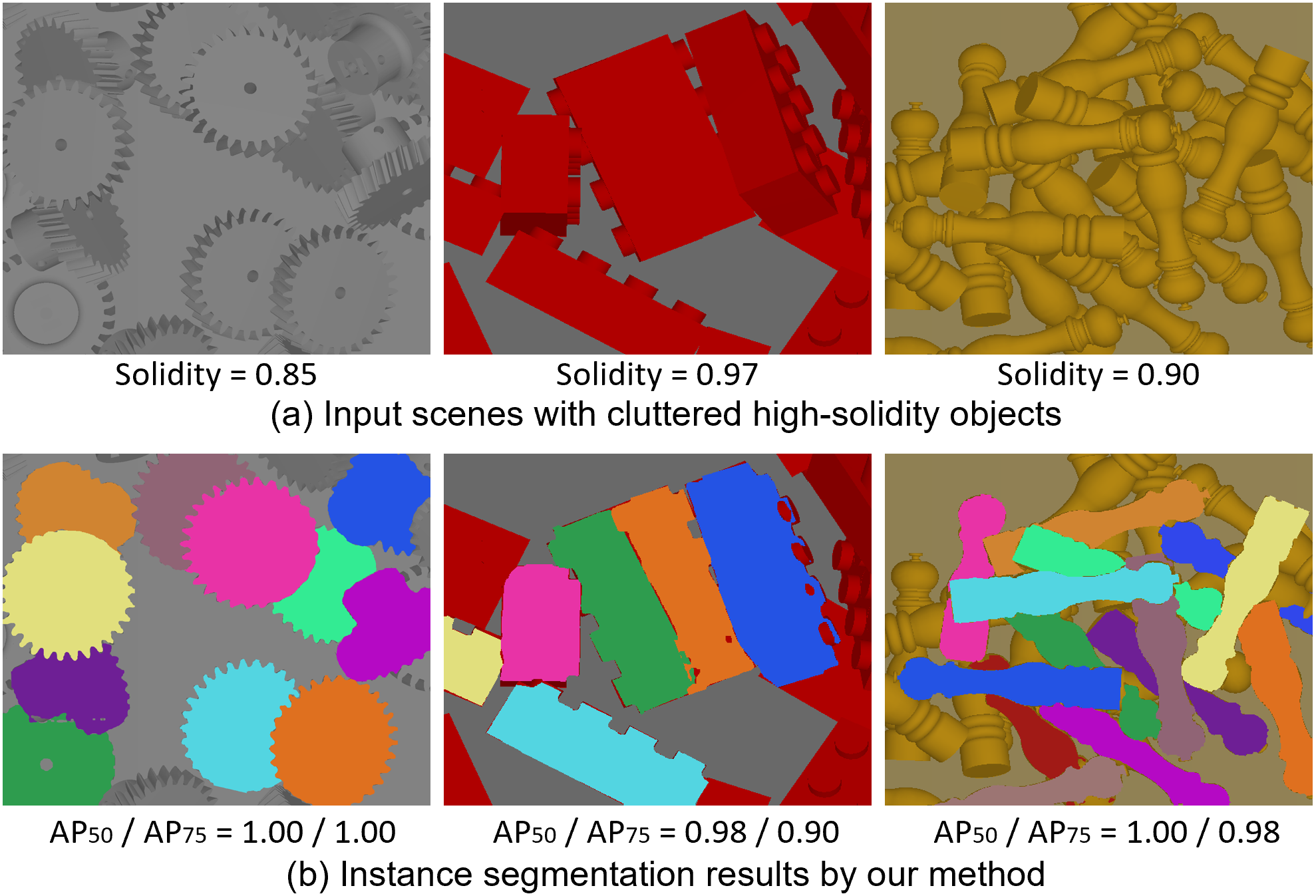}
		\vspace*{-6mm}
		\caption{Our method can be adapted to segment high-solidity objects.}
		\label{fig:highsolidity}
		\vspace*{-6mm}
	\end{figure}

	\para{Segmentation on high-solidity objects}: Besides segmenting low-solidity objects, our method can also be easily adapted to segment high-solidity objects by setting the part number $N=1$, and all other hyper-parameters are kept the same.
	Figure~\ref{fig:highsolidity} shows the results, demonstrating that our method can still achieve high precision on high-solidity objects. 
	
	\section{Conclusion}
	\label{sec:conclusion}
	
	In this work, we for the first time tackle the challenge to robustly and accurately segment low-solidity objects in cluttered industrial scenarios.
	Our key idea is to decompose a low-solidity object into multiple high-solidity parts and enhance the instance segmentation task with part awareness.
	In detail, we design a part-aware instance segmentation network to predict part-level masks and correlations, which are fed into a bidirectional aggregation module to assemble parts into object instances. 
	To guide network training, we also propose a concavity-based label decoupling scheme to generate part-level labels from instance labels. Lastly, we contribute the first segmentation dataset with various low-solidity industrial objects. 
	Our method achieves the state-of-the-art performance on these objects, and can be robustly utilized in practical bin picking by robotic grasping.

	\section{Acknowledgement}
	This work was supported by the Hong Kong Centre for Logistics Robotics, the Research Grants Council-General Research Fund (No. 14201620) and the National Natural Science Foundation of China (No. 62172218).
	
	\bibliographystyle{IEEEtran}
	\bibliography{ref}
\end{document}